\documentclass[a4paper]{article}

\usepackage{INTERSPEECH2021}
\usepackage[super]{nth}
\usepackage[ruled,vlined]{algorithm2e}

\title{Spoken Term Detection and Relevance Score Estimation\\ using Dot-Product of Pronunciation Embeddings}
\name{Jan Švec$^1$, Luboš Šmídl$^1$, Josef V. Psutka$^1$, Aleš Pražák$^1$}
\address{
  $^1$Department of Cybernetics, University of West Bohemia, Pilsen, Czech Republic}
\email{\{honzas,smidl,psutka\_j,aprazak\}@kky.zcu.cz}

\begin{document}

\maketitle
\begin{abstract}
  The paper describes a novel approach to Spoken Term Detection (STD) in large spoken archives using deep LSTM networks. The work is based on the previous approach of using Siamese neural networks for STD and naturally extends it to directly localize a spoken term and estimate its relevance score.
  The phoneme confusion network generated by a phoneme recognizer is processed by the deep LSTM network which projects each segment of the confusion network into an embedding space.
  The searched term is projected into the same embedding space using another deep LSTM network. The relevance score is then computed using a simple dot-product in the embedding space and calibrated using a sigmoid function to predict the probability of occurrence. The location of the searched term is then estimated from the sequence of output probabilities.
  The deep LSTM networks are trained in a self-supervised manner from paired recognition hypotheses on word and phoneme levels. The method is experimentally evaluated on MALACH data in English and Czech languages.
\end{abstract}

\section{Introduction}

The task of spoken term detection (STD) for large spoken archives typically employs a large vocabulary continuous speech recognition (LVCSR). By recognizing and pre-indexing the spoken utterances the in-vocabulary (IV) queries could be directly found in the word index. The handling of out-of-vocabulary (OOV) terms consists of a much wider spectrum of methods \cite{karakos2015} including recognition and indexing of sub-word units (phonemes, syllables or word fragments) \cite{Psutka2011,he2016,heerden2017}, the use of IV proxy words \cite{Zhiqiang2017,Chen2013} or the use of acoustic embeddings and similarity metrics in a vector space  \cite{Settle2016,Yuan2020}. The acoustic embeddings often play a role also in the query-by-example (QbE) task in the low-resourced setup but the idea of neural-network-based projection of the query and the utterance into a single space could be reused in the more general STD task employing the standard speech recognition models \cite{Sacchi2019,Wang2018,Kamper2016}.

In the QbE task the recurrent neural networks (RNNs) are usually used in the Siamese configuration -- two similar networks handle both the utterance and the query. The networks are often trained using the triplet loss function \cite{gundogdu2017,Zhu2018,Yuan2020}. The use of RNNs to process the signal and the query is also present in the wakeup word detection task \cite{Wang2020,Chen2015,bulin2019lstm}.

Since we are targeting the large spoken archives for which the LVCSR system exists and is used for searching the IV terms, we focused on the methods where the STD is performed using the phoneme recognizer (in a structure similar to LVCSR) to search the OOV terms. The idea is not new and we used it in a mostly heuristic search presented in \cite{Psutka2011} and subsequently we adopted the approach of Siamese networks \cite{svec2017} to robustly estimate the term relevance scores. In this work we modified the Siamese architecture presented in \cite{svec2017} with the goal to simplify the network architecture and further improve the STD performance:

\noindent \emph{STD process} -- while the original Siamese architecture was proposed to relevance score estimation only and the localization of terms was performed using the index of phoneme triplets, the proposed approach allows both to localize and score the putative hits of the searched term. The proposed method does not need any kind of DTW  \cite{He2017,gundogu2019} nor subsequence matching \cite{Wang2018}.

\noindent \emph{Network structure} -- we keep the dual structure of the network where we map the recognized sequence and searched term into an embedding space using recurrent neural networks with the same architecture \cite{Zhu2018,Settle2016}

\noindent \emph{Loss function} -- the loss function based on normalized cosine similarity \cite{He2017,Zhu2018} was replaced with a simple binary cross-entropy which allows the network outputs to be interpreted as probabilities of occurrence improving the calibration of the relevance scores \cite{Hout2014}.

\noindent \emph{Network training} -- the idea of self-supervised learning from "blindly" recognized hypotheses on word and phoneme levels was used in a similar way. This way a large amount of training data could be easily generated from large spoken archives \cite{Kamper2016,svec2017}. Moreover, such training data match exactly the speech recognizer used and the neural network could model and partially compensate the errors of the recognizer.

\noindent \emph{Pronunciation embeddings} -- both the recognition output in the form of phoneme confusion network (sausage) and the graphemic representation of the searched term are projected into the same embedding space \cite{Settle2016,Kamper2016,He2017,Yuan2020}. In this space, the probability of occurrence is computed as a dot-product of two vectors and calibrated using a simple sigmoid function.

The above-mentioned ideas like the unsupervised transformation of the utterance into the latent embedding space using wav2vec method \cite{Schneider2019}, mapping of words into the phonetic embedding space \cite{Wang2018,Sacchi2019} or the calibration of the relevance scores \cite{6707731} are broadly used. This paper presents a novel application of such methods to the STD task in an integrated seamless way.

\section{Deep LSTM for Spoken Term Detection}

The proposed network architecture reuses some ideas from \cite{svec2017} especially the mapping from the graphemic representation of a searched term and from recognized phonemes into a shared embedding space where the relevance score is easily computed. The key difference from the previous work is that the sequence of recognized phonemes is mapped to a sequence of embedded vectors of the same length while the Siamese network used the mapping to a single vector. This way the scores between the searched term and the phoneme sequence are computed on a per-phoneme basis (or per confusion network segment if using multiple phoneme hypotheses). Because the scores are computed for each phoneme we can easily determine locations of the putative hits using simple thresholding of scores. The overall relevance score of the putative hit is determined as an average of per-phoneme scores. To avoid the matches of query substrings the minimum length of the putative hit is also estimated by the network and all putative hits shorter than this minimum length are ignored.

\subsection{Network architecture}

\begin{figure}[t]
  \centering
  \includegraphics[width=1\linewidth]{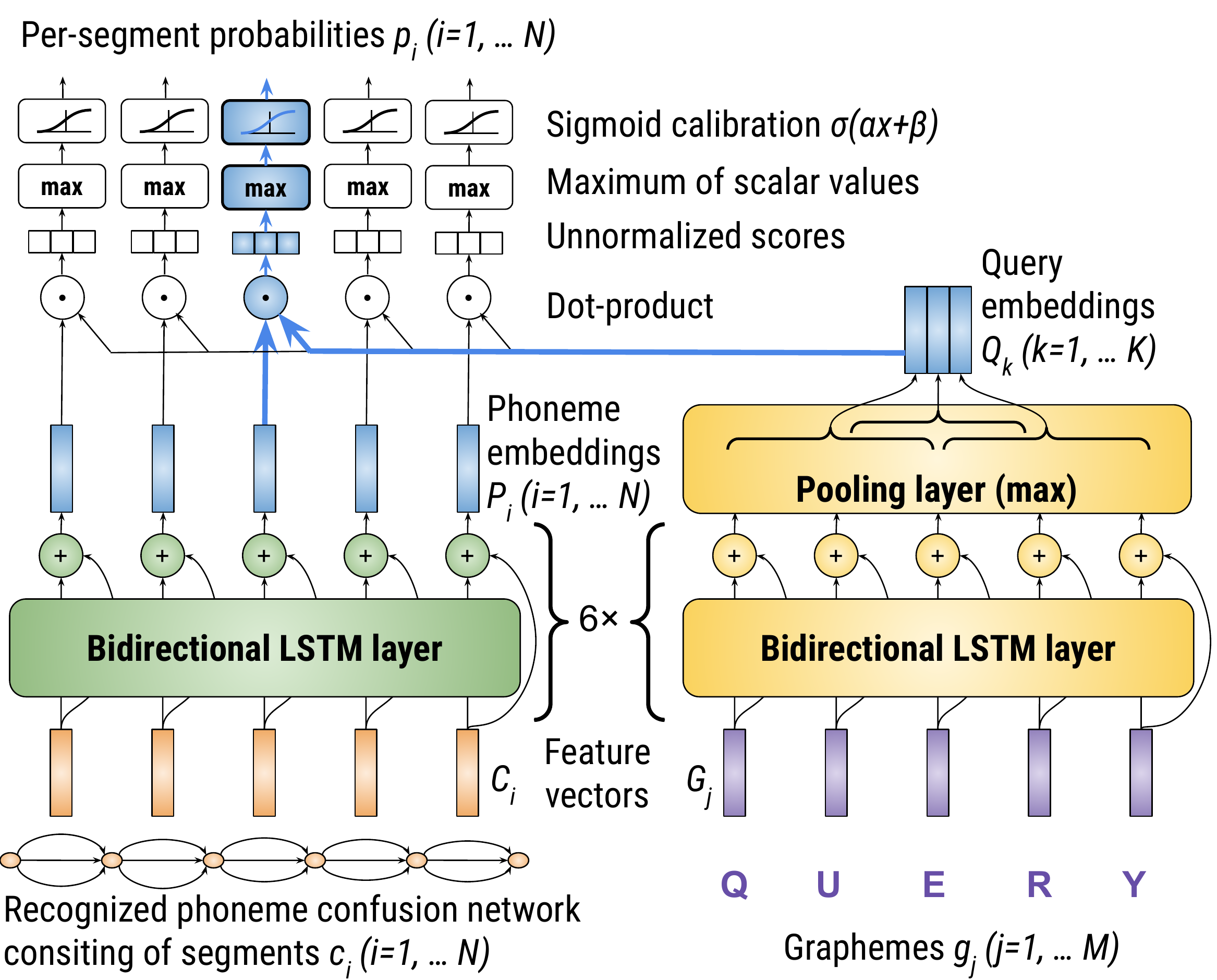}
  \caption{Deep LSTM network architecture. In blue color the scoring for a single confusion network segment is highlighted (see Eq. (\ref{eq:prob}) for details). The network for estimation of the minimal number of segments for the query is omitted for clarity.}
  \label{fig:deeplstm}
  \vspace{-1.5em}
\end{figure}

The overall network consists of two independent processing pipelines implemented using deep LSTM networks: (1) the recognition output projection and (2) the searched term projection and minimum length estimation (Fig. \ref{fig:deeplstm}). In each pipeline, we use a stack of bidirectional LSTM layers with skip connections. In our experiments, we used 6 such bidi-LSTM layers but the number of layers (and the number of trainable parameters) could be easily adjusted according to the amount of training data available. We used the dimensionality 300 (concatenated forward and backward LSTM outputs) for both processing pipelines. 

\begin{figure}[t]
  \centering
  \includegraphics[width=1\linewidth]{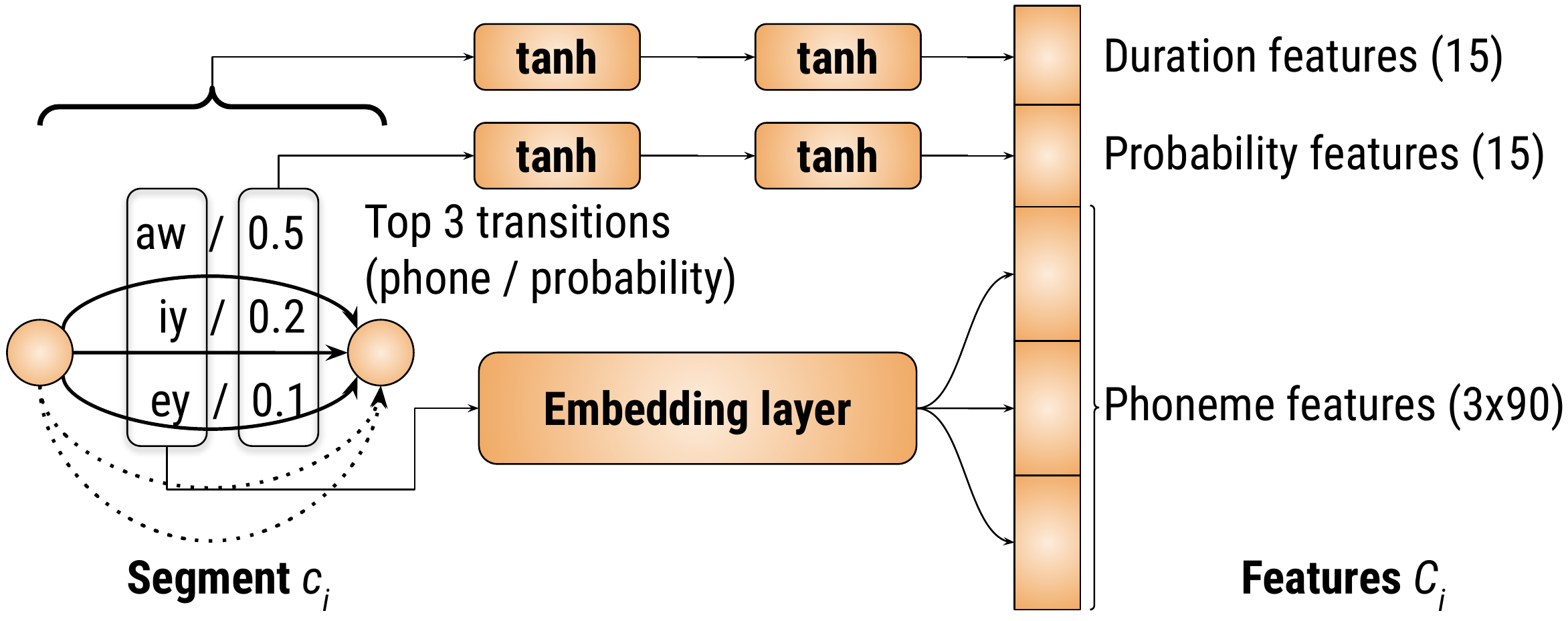}
  \caption{Extraction of features from a segment of a phoneme confusion network.}
  \label{fig:phonemeextractor}
  \vspace{-1.5em}
\end{figure}

To process the recognition output in the form of a phoneme-confusion network we use the features computed on the segments of the confusion network (i.e. on the transitions between two subsequent states). In our experiments, we used the top 3 transitions and the following features: time duration of the segment, probability of the transition, phoneme associated with the transition. If the segment contains less than 3 transitions we use padding transitions with zero probability and labeled with a special padding symbol. Alongside the padding symbol, we also use a special epsilon symbol which means transition without any phoneme associated. Such transitions are generated by the phoneme recognizer as part of the confusion network. The phonemes and special symbols are mapped to a continuous space using an embedding layer shared by all confusion network segments and by all symbols in the top 3 transitions of the segment. 
Segment duration and transition probabilities are processed using two independent stacks of two dense layers with \emph{tanh} activation function each having 15 hidden units. The dimensionality of the phoneme embedding layer is 90. This way we obtain 300 features ($15+15+3\times90$) for the phoneme processing pipeline (Fig. \ref{fig:phonemeextractor}). 

Let's denote the segments of the phoneme confusion network as $c_i, i = 1, \ldots N$ and the graphemic representation consists of graphemes $g_j, j=1, \ldots M$ (see Fig \ref{fig:deeplstm} for graphical explanation). The segments of the network are mapped to a sequence of feature vectors $C_i$ using the feature extractor described above. The graphemes are mapped using an input embedding to features $G_j$.

The feature vectors $C_i$ are mapped using the phoneme processing pipeline (6 layers of bidirectional LSTM with skip connections, green blocks in Fig. \ref{fig:deeplstm}) to the sequence of pronunciation embeddings $P_i$. For the graphemes the same architecture is used followed by the maximum pooling layer (yellow blocks in Fig. \ref{fig:deeplstm}). The output of the pooling are vectors $Q_k, k=1, \ldots K$ where the number $K$ depends on the configuration of the pooling layer (pooling size and stride). The final probability $p_i, i=1, \ldots N$ is computed as:

\begin{equation}
    p_i = \sigma\left(\alpha \cdot \max_{k=1}^K ( P_i \cdot Q_k ) + \beta\right)
    \label{eq:prob}
\end{equation}

\noindent where $\sigma(x)=\frac{1}{1+e^{-x}}$ denotes the sigmoid function and $\alpha$ and $\beta$ are trainable calibration parameters.

Since we are using the additive skip connections of the LSTM layers, the dimensionality of $C_i$, $G_j$, $P_i$, and $Q_k$ have to be the same. In our experiments, we used 300-dimensional vectors. The embedding vectors $P_i$ are independent of the query and could be pre-computed ahead of time to speed up the STD.
The pooling layer of the grapheme processing pipeline allows extracting the query embeddings $Q_k$ for different parts of the query word, for example, if $K=3$ (pool size 8 and stride 4), the $Q_1$ represents the first half of the query, $Q_2$ the middle of the query and $Q_3$ the second half. It extends the expressive power of the model without increasing the number of trainable parameters. We illustrated the matches of particular $Q_k$ for different queries in Fig. \ref{fig:std}. It could be clearly seen that the overall maximum envelope of the peak consists of three partial peaks for particular $Q_k$ vectors. By using the maximum in Eq. (\ref{eq:prob}) we do not explicitly split the target probability between different $Q_k$ vectors, we only say that on some location in the training chunk at least one of the query embeddings $Q_k$ have to match the phoneme confusion network.

The sequence of graphemic embeddings $G_i$ is also used to predict the expected minimum number of confusion network segments for the query. For this part of the network, we use a simple bidirectional LSTM with 20 units for each direction and the final outputs in each direction are combined using a dense linear layer.

\subsection{Network training}

To train the neural network we use the self-supervised scheme where the network is trained not from the ground-truth human-annotated data but from the data produced by the speech recognizer. We recognize the data using two recognition engines, one working on word level and the second one on the phoneme level. The "blind" two-level recognition allows access to a large collection of training data, even much larger than the training data for speech recognizer and the data include the typical representatives of recognition errors.

The training data consist of the phoneme-level transcriptions (represented by phoneme confusion networks) segmented into chunks containing at most $N$ segments. From the word-level representation of each chunk, the graphemic representation of the query is randomly sampled. We randomly select a single word with a probability of 0.5, two consecutive words with a probability of 0.25, etc. To avoid training from very noisy examples, we sample only words with a confusion score better than 0.95. If there is no such word, we randomly sample the recognition vocabulary to provide also negative examples during training. The selected words are joined together without space to form a query and therefore randomly simulate the OOV terms. Only queries longer than 4 graphemes and shorter than the maximum length $M=16$ are allowed.

The target value is set to 1 if the confusion segment corresponds to the sampled query, 0 otherwise. If the negative example is used, the target consists of all 0's. Then the network is trained using a standard ADAM algorithm and binary cross-entropy loss function.

The network for predicting the minimum number of confusion network segments is trained using the mean-square-error loss function. As the target value, we use the \nth{5} percentile of the number of segments seen in the training data for the query. If the query consists of multiple subsequent words, the target value is the sum of \nth{5} percentiles for each word in the query.

\subsection{STD algorithm}

\begin{figure}[t]
  \centering
  \includegraphics[width=1\linewidth]{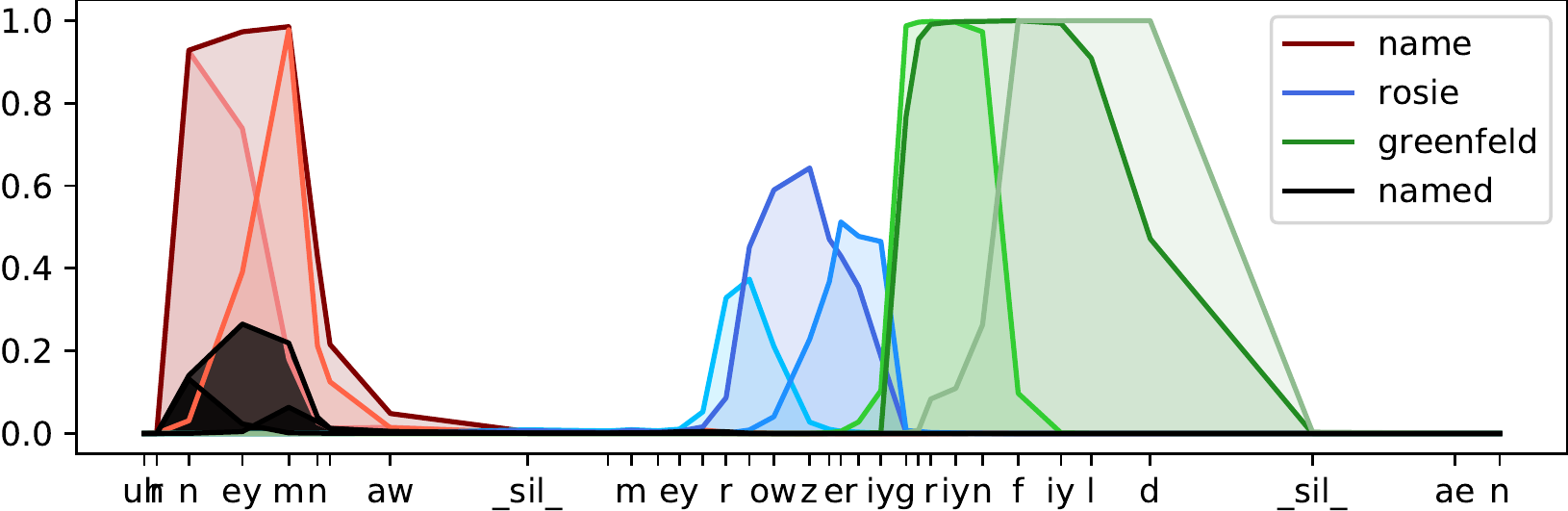}
  \caption{Per-segment STD probabilities of a selected utterance, x-axis show the time and the 1-best phoneme sequence from the confusion network. The figure highlights three words used as separate queries: \emph{name, rosie, greenfeld} while the fourth term \emph{named} is used as a counterexample to the query \emph{name}. Each query was represented as three pronunciation vectors (pooling size 8, stride 4) and the matches for each vector are shown in different shades.}
  \label{fig:std}
  \vspace{-1.5em}
\end{figure}

The spoken term detection using the deep LSTM approach consists of straightforward application of the two processing pipelines. The recognized phoneme confusion network is mapped to a sequence of embedding vectors $P_i$ just once because the embedding vectors do not depend on the searched term. Each searched term graphemic representation is mapped to a set of $K$ embedding vectors $Q_k$. The number of embedding vectors depends on the structure of the pooling layer used at the end of the query pipeline.

The sequence of embedding vectors $P_i$ from the recognized phoneme confusion network is then combined using a dot-product with the embedding vectors of the query $Q_k$ and calibrated to probabilities according to Eq. (\ref{eq:prob}).

The last step is to convert the sequence of probability scores to a set of hits with combined scores assigned. In this step, the estimate of the minimum number of confusion network segments is used. The algorithm gradually lowers the threshold from 1.0 to 0.0 and searches for peaks of probability scores above this threshold and longer than the minimum number of segments. If such a peak occurs the time span is extracted from the corresponding segments of the confusion network the combined score is computed as an arithmetic mean of probability scores forming the peak. If the peak is evaluated as a hit the corresponding probability scores are zeroed and the threshold lowering process continues. This way we avoid including the peak multiple times in the results with different thresholds.

In the experiments, we observed that filtering the probability scores with a moving average filter of window width $W$ and uniform weights $\frac{1}{W}$ improves the STD performance. The effect of this filtering is evaluated in Sec. \ref{sec:experiments}.

\section{Dataset description}

The presented method was evaluated on the data from a USC-SFI MALACH archive in two languages -- English \cite{MALACHen}
and Czech \cite{MALACHcz}. The archive
for each language was recognized using the word- and phoneme-level recognizers. The phoneme confusion networks are aligned with the word-level hypothesis and the words with a confidence score (generated by a recognition engine) higher than 0.95 were used in training. More exactly the words were used to generate the queries and the phoneme confusion networks were used entirely for training regardless of the confidence of the corresponding words.

\textbf{Speech recognition.} We followed a typical Kaldi training recipe for a time delay neural network (TDNN) acoustic model training. This recipe
supports layer-wise RBM pre-training, stochastic gradient descent training supported by GPUs and sequence-discriminative
training optimizing sMBR criterion. We applied the standard 6 layers topology (5 hidden layers, each with 2048
neurons) with a softmax layer. The output dimension was equal to the number of context-dependent states
(4521 for English, 4557 for Czech). We used features based on standard 12-dimensional Cepstral Mean Normalized (CMN) PLP
coefficients with first and second derivatives. In total, the English acoustic model was trained from 217 hours and the
Czech from 84 hours of the signal. We used our in-house real-time decoder both for the word- and phoneme recognition with trigram word- and 5-gram phoneme language model.

\begin{table}[t]
    \small
  \caption{Statistics of development and test sets \cite{svec2017}.}
  \label{tab:stats}
  \centering
  \begin{tabular}{lrrrr}
    \toprule[0.9pt]
    & \multicolumn{2}{c}{English} & \multicolumn{2}{c}{Czech} \\
    \cmidrule(l){2-3} \cmidrule(l){4-5}
    & Dev & Test & Dev & Test \\
    \cmidrule(l){1-5}
    LVCSR vocabulary & \multicolumn{2}{c}{243,699} & \multicolumn{2}{c}{252,082} \\
    \#speakers  & 10 & 10 & 10 & 10  \\
    OOV rate & 0.5\% & 3.2\% & 0.3\% & 2.6\%  \\
    LVCSR WER & 24.10 & 19.66 & 23.98 & 19.11 \\
    \#IV terms & 597 & 601 & 1680 & 1673 \\
    \#OOV terms & 31 & 6& 1145 & 948 \\
    dataset length $[$hours$]$ & 11.1 & 11.3 & 20.4 & 19.4 \\
    \bottomrule[0.9pt]
  \end{tabular}
  \vspace{-2em}
\end{table}

\section{Experimental evaluation}
\label{sec:experiments}

\begin{table}[t]
    \small
  \caption{Results on the development dataset (MTWV) using phoneme-based STD.}
  \label{tab:results}
  \centering
  \begin{tabular}{lrr}
    \cmidrule[0.9pt](l){1-3}
     & English & Czech \\
    \cmidrule(l){1-3}
     Empirical method \cite{Psutka2011}   & 0.4636 & 0.6225 \\
     Siamese neural network \cite{svec2017} & 0.5012 & 0.6547 \\
    \cmidrule(l){1-3}
     \multicolumn{3}{l}{\textbf{Deep LSTM network} (proposed method)} \\
     ~~pool size 16 ($K=1$, $W=5$) & 0.6389  & 0.7505  \\
     ~~pool size 8, stride 8 ($K=2$, $W=5$) & 0.6676 & 0.7682 \\
     ~~pool size 8, stride 4 ($K=3$, $W=5$) & \textbf{0.6703} & \textbf{0.7723} \\
     ~~~~larger filtering window ($W=7$) & 0.6567 & 0.7696 \\
     ~~~~smaller filtering window ($W=3$) & 0.6646 & 0.7714      \\
     ~~~~w/o output filtering ($W=1$) & 0.6433  & 0.7595       \\
     ~~w/o LSTM skip connections  & 0.6666 & 0.7401 \\
    \cmidrule(l){1-3}
    In-vocabulary terms & 0.6726 & 0.7755 \\
    Out-of-vocabulary terms & 0.6345 & 0.7671      \\
    \cmidrule[0.9pt](l){1-3}
  \end{tabular}
  \vspace{-1.5em}
\end{table}

The experimental evaluation was performed in the same setup as in \cite{svec2017}. To directly compare the results we reused the recognition models used in this paper. The baseline for the experiments was the empirical method from \cite{Psutka2011} and the Siamese neural network from \cite{svec2017}.
In the experiments, we compared different setups of the proposed method (see Tab. \ref{tab:results}) using the development dataset. We report the ATWV using the optimal decision threshold on development data (virtually MTWV) and the optimal decision thresholds used for STD on test data (true ATWV) \cite{Computer2013}.
Because the loss function used in network training is not necessarily correlated with the ATWV metric, we trained multiple networks using the same data and we stored all weights across the first 32 epochs. Each epoch consisted of 10k batches of 32 examples. Each example contained a chunk of $N=256$ confusion network segments. Then we used these trained networks to perform the STD on the development data and we selected the best performing network in terms of the ATWV metric (Tab. \ref{tab:results}). This network was then used in STD over the test data (Tab. \ref{tab:results-test}).

First, we show the results for different configurations of the query pooling layer. We experimented with a single global maximum pooling (pool size 16 and $K=1$ query embedding vectors), maximum pooling of the first and second half (pool size 8, stride 8 and $K=2$) and maximum pooling with three output vectors (pool size 8, stride 4 and $K=3$). The results clearly show that using more than one grapheme embedding vector brings a substantial improvement in the performance. The increase in MTWV from $K=2$ to $K=3$ is smaller than the step from $K=1$ to $K=2$.

Then, for the optimal pooling configuration, we tried different sizes of the output probability filtering window ($W$). The optimal size of the window is $W=5$ and again the results are substantially better for any $W > 1$, i.e. in comparison with the case where no filtering is used.

We also performed the experiments with a pure LSTM network without the skip connections. In this case, we used the optimal pooling and filtering window configuration ($K=3$, $W=5$). The MTWV metric shows worse performance in both languages. We also observed much slower convergention of the training process.

The last two lines in Tab. \ref{tab:results} compares the MTWV values for IV and OOV terms. This illustrates the expectation that the IV terms are easier to detect using the proposed method due to two reasons: (1) the deep LSTM network is trained from recognized IV words and (2) the language model of the phoneme recognizer (5-gram) is trained from textual transcriptions and therefore better modeling the IV words than the OOV words. 

\section{Conclusions}

\begin{table}[t]
    \small
  \caption{Results on the test dataset (ATWV).}
  \label{tab:results-test}
  \centering
  \begin{tabular}{lrr}
    \cmidrule[0.9pt](l){1-3}
    All terms & English & Czech \\
    \cmidrule(l){1-3}
    Empirical method \cite{Psutka2011}   & 0.4435 & 0.6564 \\
    Siamese neural network \cite{svec2017}  & 0.4956 & 0.6873 \\
    \textbf{Deep LSTM network} & \textbf{0.5823} & \textbf{0.7531} \\
    \cmidrule(l){1-3}
    LVCSR+Empirical method & 0.7191 & 0.6840 \\
    LVCSR+Siamese neural network   & 0.7263 & 0.7703 \\
    LVCSR+\textbf{Deep LSTM network}  & \textbf{0.7275} & \textbf{0.8283} \\
    \cmidrule[0.9pt](l){1-3}
  \end{tabular}
  \vspace{-1.5em}
\end{table}

To conclude the paper we used the trained networks with meta parameters and decision thresholds evaluated on development data (Tab.~\ref{tab:results}) to perform the STD on the test data. We present the results for both the pure phoneme-based search (top part of Tab. \ref{tab:results-test}) and for the case where the IV terms are searched using the LVCSR hypothesis and OOV terms using the proposed method (bottom part of Tab. \ref{tab:results-test}). In both cases, the ATWV metric on test data outperforms the baseline methods \cite{svec2017,Psutka2011}.

The improvement is not only from the better modeling of the underlying data and task but also from the ability of the model to distinguish between similar terms in different contexts. This is illustrated on a real part of an utterance (Fig. \ref{fig:std}) where the red and black peaks represent the per-segment probabilities of terms \emph{name} vs. \emph{named}. The probability of the second term is substantially lower despite being very similar to the first term.

The method outperforms the baseline and at the same time provides an integrated neural-network-based solution to the STD task. We introduced modifications that contribute to the improved performance: (1) the computation of calibrated per-segment probabilities and optimization of the binary cross-entropy loss, (2) representation of the query using multiple grapheme embeddings, (3) introducing the skip connections in the LSTM architecture and (4) the STD process employing gradual thresholding of output probabilities.
In future work, we would like to focus on improving this method in different ways:

\noindent \emph{Large spoken archives} -- since we can pre-compute the per-segment phoneme embeddings $P_i$, we would like to experiment with quantization of such embeddings followed by indexation. By using the quantized embeddings we will be able to directly select the set of the embeddings matching the query based on the grapheme embeddings $Q_k$ and the quantization codebook. We would like to explore it together with the decimation of phoneme embeddings $P_i$, i.e. with using one embedding for every n-th only.

\noindent \emph{Query-by-example} -- the method is suitable for jointly training the QbE system employing grapheme/audio queries and phoneme/audio representation of utterances. The only change is in replacing the phoneme confusion network feature extractor with an acoustic feature extractor.
\vspace{-0.3mm}
\section{Acknowledgements}

This research was by the Ministry of the Interior of the Czech Republic, project No. VJ01010108. Computational resources were supplied by the project "e-Infrastruktura CZ" (e-INFRA LM2018140).

\newpage{}

\bibliographystyle{IEEEtran}

\bibliography{mybib}


\end{document}